\documentclass[10pt,twocolumn,letterpaper]{article}

\usepackage{cvpr}
\usepackage{times}
\usepackage{epsfig}
\usepackage{graphicx}
\usepackage{amsmath}
\usepackage{amssymb}
\usepackage{pifont}

\usepackage[dvipsnames]{xcolor}
\usepackage{booktabs}
\usepackage{multirow}
\usepackage[skip=0.2\baselineskip,font=small]{caption}

\definecolor{sadness}{RGB}{74, 65, 42}

\newcommand{\proxylessnas}[0]{ProxylessNAS}
\newcommand{\mobilenetv}[1]{MobileNetV#1}
\newcommand{\mnasnet}[0]{MnasNet}

\usepackage[pagebackref=true,breaklinks=true,letterpaper=true,colorlinks,bookmarks=false,citecolor=Green]{hyperref}

\cvprfinalcopy %

\def\cvprPaperID{9997} %

\ifcvprfinal\fi

\begin{document}

\title{Can weight sharing outperform random architecture search? An investigation with TuNAS}

\author{Gabriel Bender, Hanxiao Liu\\
Google Research, contributed equally\\
{\tt\small gbender@google.com, hanxiaol@google.com}
\and 
Bo Chen\\
Google Research\\
{\tt\small bochen@google.com}
\and 
Grace Chu\\
Google Research\\
{\tt\small cxy@google.com}
\and 
Shuyang Cheng\\
Waymo\\
{\tt\small shuyangcheng@waymo.com}
\and 
Pieter-Jan Kindermans\\
Google Research \\
{\tt\small pikinder@google.com}
\and
Quoc Le\\
Google Research \\
{\tt\small qvl@google.com}
}

\maketitle

\begin{abstract}

Efficient Neural Architecture Search methods based on weight sharing have shown good promise in democratizing Neural Architecture Search for computer vision models. There is, however, an ongoing debate whether these efficient methods are significantly better than random search. Here we perform a thorough comparison between efficient and random search methods on a family of progressively larger and more challenging search spaces for image classification and detection on ImageNet and COCO. While the efficacies of both methods are problem-dependent, our experiments demonstrate that there are large, realistic tasks where efficient search methods can provide substantial gains over random search. In addition, we propose and evaluate techniques which improve the quality of searched architectures and reduce the need for manual hyper-parameter tuning.
\footnote{Source code and experiment data are available at {\scriptsize \url{https://github.com/google-research/google-research/tree/master/tunas}}}

\end{abstract}

\section{Introduction}

Neural Architecture Search (NAS) tries to find network architectures with excellent accuracy-latency tradeoffs. While the resource costs of early approaches \cite{zoph2018learning,real2019regularized} were prohibitively expensive for many, recent \emph{efficient architecture search methods} based on weight sharing promise to reduce the costs of architecture search experiments by multiple orders of magnitude.  \cite{pham2018efficient,bender2018understanding,liu2018darts,cai2018proxylessnas,xie2018snas}

The effectiveness of these efficient NAS approaches has been questioned by recent studies (e.g., \cite{li2019random,sciuto2019evaluating}) presenting experimental results where efficient architecture search methods did not always outperform random search baselines. Furthermore, even when gains were reported, they were often modest. However, most existing results come with limitations. First: negative results may simply indicate that existing algorithms are challenging to implement and tune. Second:
most negative results focus on fairly small datasets such as CIFAR-10 or PTB, and some are obtained on heavily restricted search spaces.
With those caveats in mind,
it is possible that efficient NAS methods work well only on specific search spaces and problems. But even so, they can still be useful if those problems are of high practical value.
For this reason we focus on the following: ``\emph{Can efficient neural architecture search be made to work reliably on large realistic search spaces for realistic problems?}''

\begin{figure}[t]
    \centering
    \includegraphics[width=0.725\linewidth]{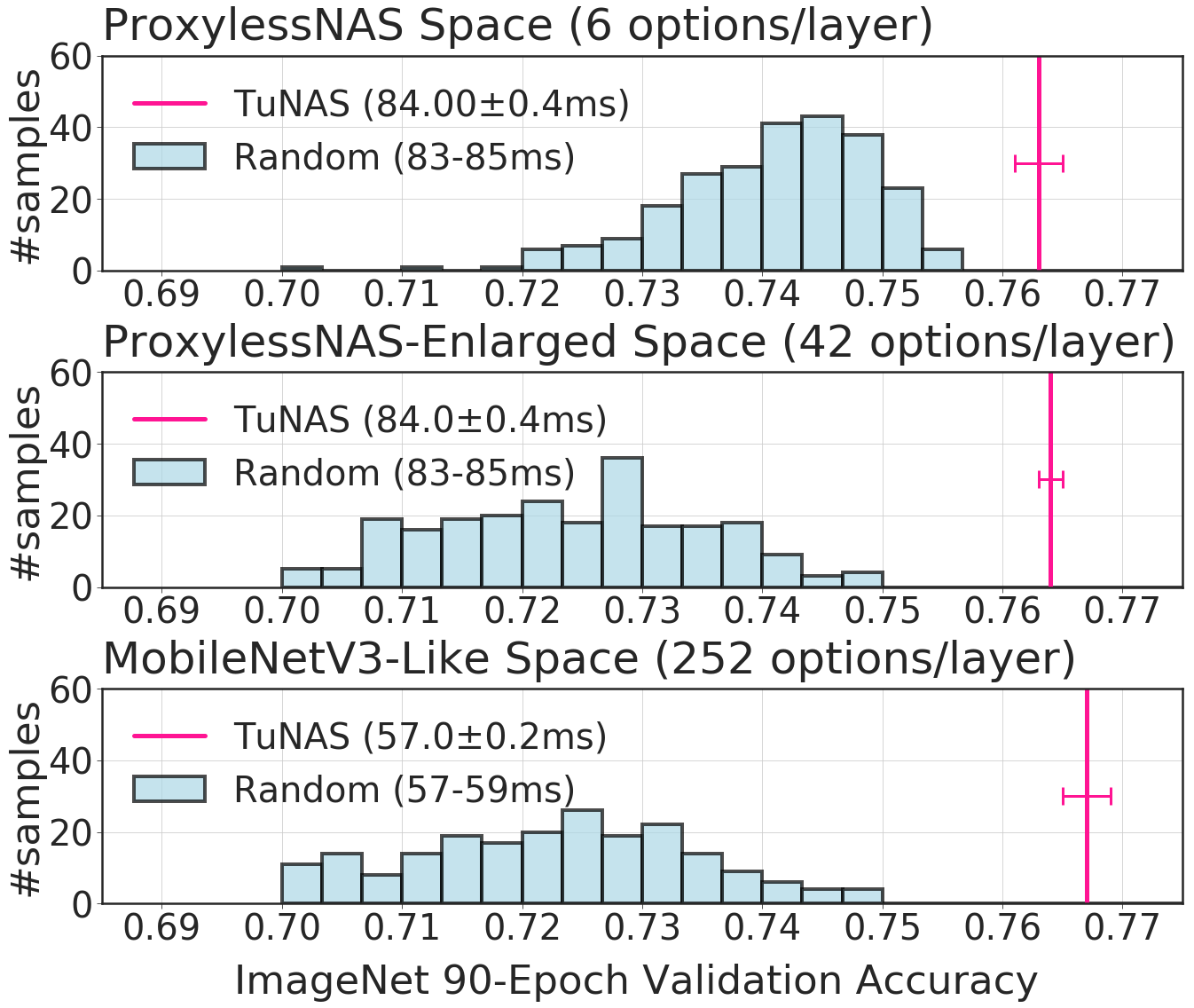}
    \caption{Validation accuracies of 250 random architectures (blue bars) vs. 5 independent runs of our TuNAS search algorithm (pink lines) for three different search spaces. Rejection sampling is used to ensure the latencies of random architectures are comparable to those of searched architectures.}
    \label{fig:baseline-vs-random-histogram}
\end{figure}

When comparing against simpler algorithms such as random search, we must consider not just \emph{explicit} costs, such as the time needed to run a single search, but also \emph{implicit costs}. For example, many models operate under hard latency constraints (e.g., a model running on a self-driving car in real time). However, the reward functions used by MnasNet \cite{tan2019mnasnet} and ProxylessNAS \cite{cai2018proxylessnas} require us to run multiple searches with different hyper-parameters to match a given latency target.\footnote{In our experiments we typically needed to run 7 searches to achieve the desired latency.} Larger models cannot be deployed, while smaller models have sub-optimal accuracies.

We present TuNAS, a new easy-to-tune and scalable implementation of efficient NAS with weight sharing, and use it to investigate the questions outlined above:
\begin{itemize}
    \item We investigate two existing models from the NAS literature, and find that while some of their gains are likely due to better training hyper-parameters, some are due to real improvements in the network architectures.
    \item We use TuNAS to study the effectiveness of weight sharing on the ProxylessNAS search space for ImageNet.
    Prior work \cite{cai2018proxylessnas} demonstrated that an efficient NAS algorithm could find state-of-the-art image classification architectures on this space, but left open the question of whether a simple random search baseline could find equally good results.
    We find that efficient NAS can significantly outperform random search on the ProxylessNAS search space on ImageNet.
    \item %
    We further evaluate our implementation on two new and even larger search spaces. We find that (i) TuNAS continues to find high-quality architectures, and (ii) the gap between TuNAS and random search increases significantly on these new search spaces. 
    \item We demonstrate that when weight sharing is implemented carefully, the same algorithm can be used across different classification search spaces \emph{and} across domains (classification and object detection).
    \item We propose and evaluate two new techniques, which we call \emph{op} and \emph{filter warmup}, for better training the shared model weights used during a search. These techniques improve the quality of architectures found by our algorithm, sometimes by a large margin.
    \item We propose a novel RL reward function which allows us to precisely control the latency of the final architecture returned by the search, significantly reducing the need for additional hyper-parameter tuning. %
\end{itemize}

\section{Related work}

\noindent \textbf{Neural Architecture Search} was proposed as a way to automate and optimize the design process for network architectures. Existing methods have achieved impressive results on image classification \cite{zoph2018learning,xie2019exploring,tan2019efficientnet}, object detection \cite{ghiasi2019fpn,chen2019detnas}, segmentation \cite{liu2019auto}, and video understanding \cite{pier2019tiny}. Early methods were based on Reinforcement Learning~\cite{zoph2016neural,baker2016designing}, Genetic Algorithms~\cite{real2017large,liu2017hierarchical,real2019regularized} or Bayesian Optimization~\cite{kandasamy2018neural}. As these methods require a large amount of compute to achieve good results, recent works focus on addressing this requirement \cite{baker2017accelerating,negrinho2017deeparchitect,luo2018neural,Liu_2018_ECCV}. In many of these methods, a single supernetwork is trained which encompasses all possible options in the search space \cite{brock2018smash,pham2018efficient,bender2018understanding,cai2018proxylessnas,liu2018darts,xie2018snas,cui2019fast}. A single path within the supernetwork corresponds to an architecture in the search space. With this scheme, weight sharing naturally occurs between the architectures. 

Within the framework of efficient search methods based on weight sharing, our work is closely related to ENAS \cite{pham2018efficient}, DARTS \cite{liu2018darts}, SNAS \cite{xie2018snas} and especially ProxylessNAS \cite{cai2018proxylessnas} in terms of optimizing the quality-latency tradeoff of architectures. Different from ProxylessNAS, our method is able to handle substantially larger and more difficult search spaces with less prior knowledge (\eg, hand-engineered output filter sizes) and achieve improved performance.

\vspace{1em}
\noindent \textbf{Multi-Objective Search.}
An important strength of Neural Architecture Search is that it can cope with an objective function beyond pure accuracy. Recently, Neural Architecture Search has been used intensively to find architectures that have better tradeoff between accuracy and latency~\cite{tan2019mnasnet,wu2019fbnet,Dai_2019_CVPR,cai2018proxylessnas,stamoulis2019single}, FLOPS~\cite{tan2019efficientnet}, power consumption~\cite{hsu2018monas}, and memory usage~\cite{fedorov2019sparse}.
We also focus on resource-aware NAS because (i) finding architectures with good trade-offs between accuracy and latency is valuable in practice, and (ii) constrained optimization may be more challenging than unconstrained optimization, which makes it ideal for a stress-test of efficient NAS algorithms.
With this in mind, we make use of and extend the search spaces proposed by MnasNet~\cite{tan2019mnasnet}, ProxylessNAS~\cite{cai2018proxylessnas}, MobileNetV2~\cite{sandler2018mobilenetv2} and MobileNetV3~\cite{howard2019searching}, which are close or among the state-of-the-art networks for mobile settings.

\vspace{1em}
\noindent \textbf{Random Search vs Efficient Search Methods.} The use of more complicated search methods in Neural Architecture Search has been challenged~\cite{sciuto2019evaluating,li2019random}. In a nutshell, these studies find that for certain problems and search spaces, random search performs close to or just as well as more complicated search methods. These studies, however, mainly focused on relatively small tasks (such as CIFAR-10) and accuracy-only optimization.
Our focus is on larger and more challenging searches which incorporate latency constraints. In these more realistic settings, %
efficient architecture search significantly outperforms random search.

\begin{table*}[]
\footnotesize
\centering
\begin{tabular}{@{}c|cccccccc@{}}
\toprule
\multirow{2}{*}{Search Space} & \multirow{2}{*}{Cardinality} & \multirow{2}{*}{Ref Model} & \multirow{2}{*}{Our Search} &  \multicolumn{4}{c}{Random Search} & \multirow{2}{*}{Simulated} \\ \cmidrule{5-8}
 &  &  &  & N=1 & N = 20 & N = 50 & N = 250 & Latency (ms) \\ \midrule
ProxylessNAS & $\sim$10$^{21}$ & 76.2 & $76.3 \pm 0.2$ & 74.1 $\pm$ 0.8 & 75.4 $\pm$ 0.1 & 75.4 $\pm$ 0.2 & 75.6 & 83-85 \\
ProxylessNAS-Enlarged & $\sim$10$^{28}$ & 76.2 & $76.4 \pm 0.1$ & 72.1 $\pm$ 1.5 & 74.4 $\pm$ 0.5 & 74.6 $\pm$ 0.3 & 74.8 & 83-85 \\
MobileNetV3-Like & $\sim$10$^{43}$ & 76.5 & $76.6 \pm 0.1$ & 71.7 $\pm$ 1.7 & 74.1 $\pm$ 0.6 & 74.6 $\pm$ 0.3 & 74.9 & 57-59 \\ \bottomrule
\end{tabular}
\caption{Comparison between reference models proposed in previous work (``Ref Model''), random search baselines in our search spaces (``Random Search''), and searched models found by TuNAS (``Our Search''). We report \emph{validation} accuracies on ImageNet after 90 epochs of training.
Cardinality refers to (an upper bound of) the number of unique architectures in the search space. The reference model for the ProxylessNAS and ProxylessNAS-Enlarged search spaces is our reproduction of the ProxylessNAS mobile CPU model~\cite{cai2018proxylessnas}. The reference model for the MobileNetV3-Like search space is our reproduction of MobileNetV3~\cite{howard2019searching}. Mean and variance for Random Search are reported over 5 repeats for N=20 and N=50, and 250 repeats for N=1.} 
\label{tab:search-space-progression}
\end{table*}

\begin{table*}[th!]
\footnotesize
\centering
\begin{tabular}{@{}c|cccccc@{}}
\toprule
\multirow{2}{*}{Search Space} &
\multirow{2}{*}{Built Around} & \multirow{2}{*}{\begin{tabular}[c]{@{}c@{}}Base Filter Sizes \\[0.6em] ($c_i$'s for each layer)\end{tabular}} & \multicolumn{4}{c}{Typical Choices within an Inverted Bottleneck Layer} \\ \cmidrule(l){3-7} 
 & & & Expansion Ratio & Kernel & Output filter size & SE \\ \cmidrule(r){1-7}
ProxylessNAS & \mobilenetv{2} & ProxylessNAS \cite{cai2018proxylessnas} & \{3, 6\} & \{3, 5, 7\} & $c_i$ & \ding{55} \\
\\[-.7em]
ProxylessNAS-Enlarged & \mobilenetv{2} & $\times 2$ when stride $=2$ & \{3, 6\} & \{3, 5, 7\} & $c_i \times \left\{\frac{1}{2}, \frac{5}{8}, \frac{3}{4}, 1, \frac{5}{4}, \frac{3}{2}, 2\right\}$ & \ding{55} \\
\\[-.7em]
MobileNetV3-Like & \mobilenetv{3} & $\times 2$ when stride $=2$ & \{1, 2, 3, 4, 5, 6\} & \{3, 5, 7\} & $c_i \times \left\{\frac{1}{2}, \frac{5}{8}, \frac{3}{4}, 1, \frac{5}{4}, \frac{3}{2}, 2\right\}$ & \{\ding{55}, \ding{51}\} \\
\bottomrule
\end{tabular}
\caption{Search spaces we use to evaluate our method. The first two are built around \mobilenetv{2}. The third uses the combination of ReLU and SiLU/Swish~\cite{ramachandran2017searching,elfwing2018sigmoid,hendrycks2016gaussian} activations and the new model head from \mobilenetv{3}. We use a target inference time of 84ms for the first two (to compare against \proxylessnas{} and \mnasnet{}) and 57ms for the third search space (to compare against \mobilenetv{3}).}
\label{tab:search-space-definitions}
\end{table*}

\section{Search Spaces} \label{sec:search-spaces}

Our goal is to develop a NAS method that can reliably find high quality models at a specific inference cost across multiple search spaces.
We next present three progressively larger search spaces and show that they are non-trivial: they contain known good reference models\footnote{In Table \ref{table:baseline-model-repro} we present our reproductions to the published numbers. In all cases our reproductions are at least as accurate as the published results.} that clearly outperform models found by random search, as shown in Table~\ref{tab:search-space-progression}. The same table shows that for the larger of these search spaces, the gap between known good models and random search baselines widens. Although architecture search becomes more difficult, it can also be more beneficial.

\subsection{Search Space Definitions}

Details of the three search spaces are summarized in Table~\ref{tab:search-space-definitions}. Motivations for each of them are outlined below.

\vspace{1em}
\noindent \textbf{ProxylessNAS Space.} The first and the smallest search space is a reproduction of the one used in \proxylessnas~\cite{cai2018proxylessnas},
an efficient architecture search method that reports strong results on mobile CPUs.
It consists of a stack of inverted bottleneck layers,
where the expansion ratio and the depthwise kernel size for each layer are searchable.
The search space is built around MobileNetV2 \cite{sandler2018mobilenetv2}, except that the output filter sizes for all the layers are handcrafted to be similar to those found by \mnasnet{} \cite{tan2019mnasnet}.

\vspace{1em}
\noindent \textbf{ProxylessNAS-Enlarged Space.}
While earlier convolutional architectures such as VGG \cite{simonyan2014very} used the heuristic of doubling the number of filters every time the feature map width and height were halved, more recent models \cite{sandler2018mobilenetv2,tan2019mnasnet,cai2018proxylessnas} obtain strong performance using more carefully tuned filter sizes. Our experimental evaluation (Table \ref{tab:output-filter-sizes}) demonstrates that these carefully tuned filter sizes are in fact important for obtaining good accuracy/latency tradeoffs.

While output filter sizes are something that we ideally should be able to search for automatically, the original ProxylessNAS search space used manually tuned filter sizes built around those discovered by an earlier and more expensive search algorithm \cite{tan2019mnasnet}. To understand whether this restriction can be lifted, we explore an extension of the ProxylessNAS search space %
to automatically search over the number of output filters in each layer of the network. %

Specifically,
we define a list of possible output filter sizes for each layer in our search space by multiplying an integer-valued \emph{base filter size} by a predefined set of multipliers $\left\{\frac{1}{2}, \frac{5}{8}, \frac{3}{4}, 1, \frac{5}{4}, \frac{3}{2}, 2\right\}$ and rounding to a multiple of 8.\footnote{For performance reasons, working with multiples of 8 was recommended for our target inference hardware.} The base filter size is 16 for the first layer of the network, and is doubled whenever we start a new block. If two layers of the network are connected via a residual connection, we constrain them to have the same output filter size. %

\vspace{1em}
\noindent \textbf{MobileNetV3-Like Space.}
Our largest search space is inspired by MobileNetV3 \cite{howard2019searching}.
Different from the previous spaces,
models in this space utilize the SiLU/Swish~\cite{ramachandran2017searching,elfwing2018sigmoid,hendrycks2016gaussian} activation function~\cite{ramachandran2017searching} and a compact head \cite{howard2019searching}.
The search space is also much larger than the previous two. First: inverted bottleneck expansion ratios can be selected from the set $\{1, 2, 3, 4, 5, 6\}$, compared with $\{3, 6\}$ in other search spaces. Second: we optionally allow a Squeeze-and-Excite module \cite{hu2018squeeze} to be added to each inverted bottleneck. Output filter sizes are searched; the choices follow the same heuristic used in the ProxylessNAS-Enlarged space.

\subsection{Measuring Search Algorithm Effectiveness}
\label{sec:difficulty}

We measure the effectiveness of our NAS algorithm on a given search space in two different ways.

\vspace{1em}
\noindent \textbf{Reference Models.} Can our algorithm match or exceed the quality of known good architectures within the space? For example, when evaluating the effectiveness of our algorithm on the ProxylessNAS-Enlarged space, we compare against MobileNetV2 \cite{sandler2018mobilenetv2} (a hand-tuned model), MnasNet \cite{tan2019mnasnet} (a model obtained from a more expensive NAS algorithm where thousands of candidate architectures were trained from scratch), and ProxylessNAS-Mobile \cite{cai2018proxylessnas} (a model obtained using a NAS algorithm similar to ours, which we use to verify our setup).

\vspace{1em}
\noindent \textbf{Random Search.} Can our algorithm provide better results in less time than random search without weight sharing, an easy-to-implement heuristic which is widely used in practice? In industry settings, it is common to target a specific latency $T_0$; slower models cannot be deployed while faster models typically have sub-optimal accuracies~\cite{tan2019efficientnet}. However, in practice only a small percentage of models are actually close to this inference target.%

To make the baseline interesting in this realistic setup,
we perform rejection sampling in the range of $T_0 \pm$1ms to obtain $N$ random models. These models are then trained for 90 epochs and validated. The model with the best result on the validation set is subsequently trained for 360 epochs and evaluated on the test set, analogous to our searched models for final evaluation. Note the cost of random search with $N=15$ to $30$ is comparable with the cost of a single run of our efficient search algorithm (Appendix \ref{sec:random-search-cost}).

Besides the comparisons discussed above, the complexity of our search spaces can be quantified using several other metrics, which we report in Table \ref{tab:search-space-progression}. A clear progression in the task difficulties can be observed as we move from the smallest ProxylessNAS search space to the largest MobileNetV3-Like search space.

\section{TuNAS}

TuNAS uses a reinforcement learning algorithm with weight sharing to perform architecture searches. Our algorithm is similar to ProxylessNAS \cite{cai2018proxylessnas} and ENAS \cite{pham2018efficient}, but contains changes to improve robustness and scalability and reduce the need for manual hyper-parameter tuning.

A search space is represented as a set of categorical decisions which control different aspects of the network architecture. For example, a single categorical decision might control whether we use a 3$\times$3, 5$\times$5, or 7$\times$7 convolution at a particular position in the network. An architecture is an assignment of values to these categorical decisions.

During a search, we learn a policy $\pi$, a probability distribution from which we can sample high quality architectures. Formally, $\pi$ is defined as a collection of independent multinomial variables, one for each of the decisions in our search space. We also learn a set of shared weights $W$, which are used to efficiently estimate the quality of candidate architectures in our search space.

We alternate between learning the shared weights $W$ using gradient descent and learning the policy $\pi$ using REINFORCE \cite{williams1992simple}. At each step, we first sample a network architecture $\alpha \sim \pi$. Next, we use the shared weights to estimate the quality $Q(\alpha)$ of the sampled architecture using a single batch of examples from the validation set. We then estimate the inference time of the sampled architecture $T(\alpha)$. The accuracy $Q(\alpha)$ and inference time $T(\alpha)$ jointly determine the reward $r(\alpha)$ which is used to update the policy $\pi$ using REINFORCE.\footnote{We set the learning rate of the RL controller to 0 during the first 25\% of training. This allows us to learn a good set of shared model weights before the RL controller kicks in. Details are provided in Appendix \ref{sec:architecture-search-setup}.} Finally, we update the shared model weights $W$ by computing a gradient update w.r.t. the architecture $\alpha$ on a batch of examples from the training set.

The above process is repeated over and over until the search completes. At the end of the search, the final architecture is obtained by independently selecting the most probable value for each categorical decision in $\pi$.

\subsection{Weight Sharing} \label{sec:aggressive-weight-sharing}
To amortize the cost of an architecture search, NAS algorithms based on weight sharing (e.g., \cite{bender2018understanding,pham2018efficient,liu2018darts,cai2018proxylessnas}) train a large network -- a \emph{one-shot model} -- with many redundant operations, most of which will be removed at the end of the search. In ProxylessNAS \cite{cai2018proxylessnas}, for instance, we must select between three possible kernel sizes (3, 5, or 7) and two possible expansion factors (3 or 6), giving us $3 \times 2 = 6$ possible combinations. In ProxylessNAS, each of these six combinations will have its own \emph{path through the network}: its own set of trainable weights and operations which are not shared with any other path. At each step, we randomly select one of the six paths and update the shared model weights and RL controller using \emph{just} the weights from the selected path.

While this approach works well when the number of paths is small, the size of the one-shot model will quickly blow up once we add more primitives to the search. For instance, the number of unique inverted bottleneck configurations per layer can be as large as $6 \times 3 \times 7 \times 2=252$ in our MobileNetV3-Like space, in contrast to $2 \times 3 = 6$ options in the ProxylessNAS space (Table \ref{tab:search-space-definitions}). As a result,
the search process cannot be carried out efficiently because each inverted bottleneck will only be trained a small fraction of time ($1/252$ under a uniform policy).

\begin{figure}
    \centering
    \includegraphics[width=.6\linewidth]{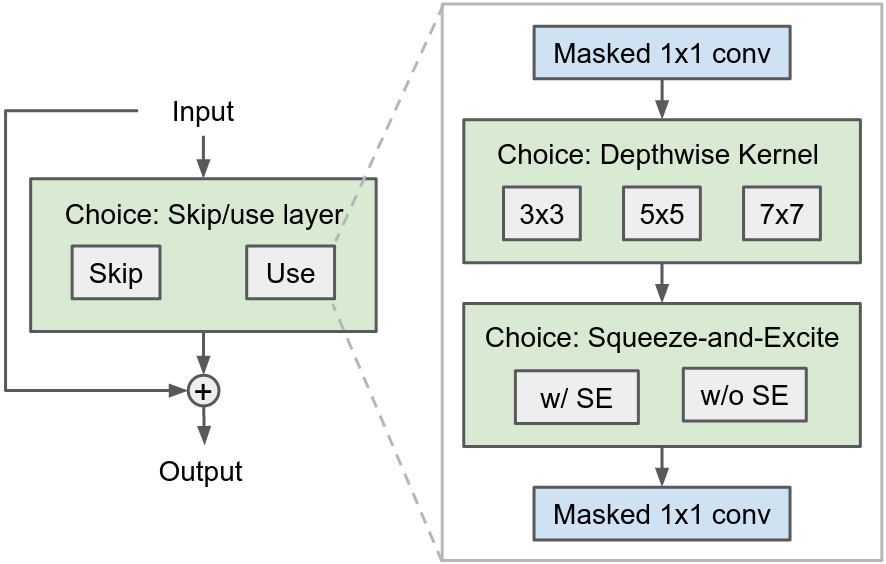}
    \caption{Illustration of our aggressive weight sharing scheme for inverted bottleneck layers. Each choice block is associated with a decision to be made based the RL policy. The expansion ratios and output filter sizes for the $1\times1$ convolutions are learned through a channel masking mechanism.}
    \label{fig:proxylessnas-collapsed}
    \vspace{-1em}
\end{figure}

\vspace{1em}
\noindent \textbf{Operation Collapsing.}
Instead of using a separate set of weights for each possible combination of choices within an inverted bottleneck, we share (``collapse'') operations and weights in order to ensure that each trainable weight gets a sufficient gradient signal. The approach is illustrated in Figure \ref{fig:proxylessnas-collapsed}. For example, while ProxylessNAS uses different $1\times1$ convolutions for each possible depthwise kernel within an inverted bottleneck, we reuse the same $1 \times 1$ convolutions regardless of which depthwise kernel is selected.

\vspace{1em}
\noindent \textbf{Channel Masking.}
Complementary to operation collapsing, we also share parameters between convolutions with different numbers of input/output channels. The idea is to create only a single convolutional kernel with the largest possible number of channels. We simulate smaller channel sizes by retaining only the first $N$ input (or output) channels, and zeroing out the remaining ones. This allows us to efficiently search for both expansion factors and output filter sizes in an inverted bottleneck, as learning these is reduced to learning multinomial distributions over the masks.

\subsection{Targeting a Specific Latency}
\label{sec:abs-reward}

For many practical applications (e.g., real-time image perception), we want the best possible model that runs within a fixed number of milliseconds. However, we found that with the existing RL reward functions used by \mnasnet{} \cite{tan2019mnasnet} and \proxylessnas{} \cite{cai2018proxylessnas}, we frequently had to retune our search hyper-parameters in order to find the best models under a given latency target. This extra retuning step multiplied resource costs by 7$\times$ in many of our experiments. We now explain why, and propose a new reward function to address the issue.

\vspace{1em}
\noindent \textbf{Soft Exponential Reward Function.} In previous work, Tan et al. \cite{tan2019mnasnet} proposed a parameterized RL reward function to find architectures with good accuracy/time tradeoffs, and evaluated two instantiations of this function. In the first instantiation (later adopted by ProxylessNAS~\cite{cai2018proxylessnas}), they maximize the reward

\vspace{-0.9em}
\begin{equation*}
    r(\alpha) = Q(\alpha) \times (T(\alpha) / T_0)^\beta
\end{equation*}
\vspace{-1.4em}

\noindent where $Q(\alpha)$ indicates the quality (accuracy) of a candidate architecture $\alpha$, $T(\alpha)$ is its inference time, $T_0$ is a problem-dependent \emph{inference time target}, and $\beta < 0$ is the \emph{cost exponent}, a tunable hyper-parameter of the setup. Since $\beta < 0$, this reward function is maximized when the model quality $Q(\alpha)$ is large and the inference time $T(\alpha)$ is small.

However, to find the best possible model whose inference time is less than $T_0$, we must perform a hyper-parameter sweep over $\beta$. If $\beta$ is too small, the inference constraint will effectively be ignored. If $\beta$ is too large, we will end up with models that have low latencies but sub-optimal accuracies. To make matters worse, we found that changing the search space or search algorithm details required us to retune the value of $\beta$, increasing search experiment costs by 7$\times$ in practice.

\begin{figure}[tb]
\centering
\includegraphics[width=1.\linewidth]{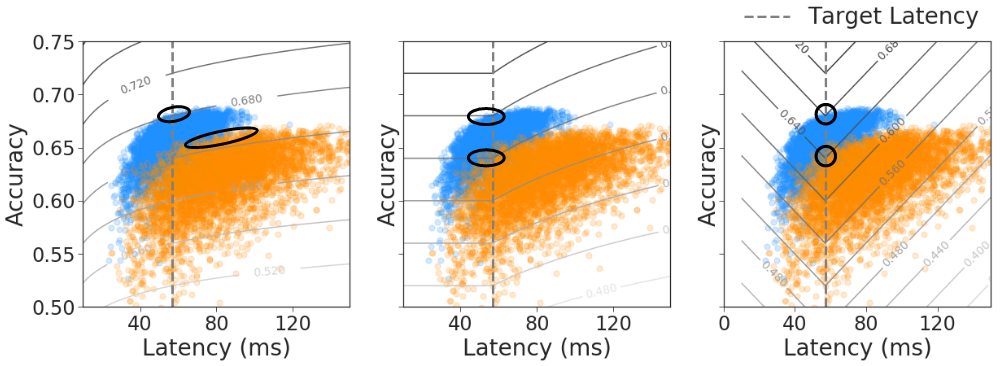}
\caption{Contours of different reward functions and their interactions with the frontiers.
The blue and the orange denote the frontiers of two search spaces with different accuracy-latency trade-offs. \emph{\textbf{Left}}: Soft exponential reward function. \emph{\textbf{Center}}: Hard exponential reward function. \emph{\textbf{Right}}: Our absolute reward function. Regions in black indicate architectures with the highest reward.}
\label{fig:reward-contours}
\end{figure}

Figure \ref{fig:reward-contours} shows a geometric intuition for this behavior. Each contour line in the plot represents a set of possible tradeoffs between model quality and latency which yield the same final reward. Our goal is to try to find an architecture with the highest possible reward, corresponding to a contour line that is as far to the top left as possible. However, the reward must correspond to a viable architecture in the search space, which means the contour must intersect the population's accuracy-latency frontier (circled in black).

For the soft exponential reward, the figure suggests that a small shift in the population (e.g., due to a change in the training setup or search space) can significantly alter the optimal latency. This explains why the same value of $\beta$ can lead to different latencies in different searches. Both the hard exponential reward function and the proposed absolute reward function, which we will discuss next,  are more stable, thanks to the ``sharp'' change points in their contours.

\vspace{1em}
\noindent \textbf{Hard Exponential Reward Function.}
A second instantiation of the \mnasnet{} reward function \cite{tan2019mnasnet} penalizes models whose inference times $T(\alpha)$ are above $T_0$ but does not reward models whose inference times are less than $T_0$:

\vspace{-1.8em}
\begin{equation}
    r(\alpha) = \begin{cases}
        Q(\alpha), & \text{if } T(\alpha) \leq T_0 \\
        Q(\alpha) \times (T(\alpha) / T_0)^\beta, & \text{if } T(\alpha) > T_0
    \end{cases}
\end{equation}
\vspace{-1.3em}

\noindent At first glance, we might expect that an RL controller using this reward would always favor models with higher accuracies, provided that their inference times do not exceed $T_0$. However, this is not the case in our experiments. The reason is that the RL controller does not select a \emph{single} point on the Pareto frontier. Rather, it learns a probability distribution over points. If the cost exponent $\beta$ is too large, the controller will become \emph{risk-adverse} preferring to sample architectures whose latencies are significantly below the target $T_0$. This allows it to minimize the probability of accidentally sampling architectures whose times exceed the target and incurring large penalties. Empirically, we found that if we made the cost penalty $\beta$ too large, the controller would sample architectures with inference times close to 75ms, even though the target inference time was closer to 85ms.%

\vspace{1em}
\noindent \textbf{Our Solution: Absolute Reward Function.}
We propose a new reward function which can find good architectures whose inference times are close to a user-specified target $T_0$ and is robust to the exact values of hyper-parameters. The key idea is to add to our reward function a prior that larger models are typically more accurate. Instead of just asking the search to identify models with inference times less than $T_0$ (as in previous work), we explicitly search for the best possible models whose inference times are close to $T_0$.
This implies a constrained optimization problem:

\vspace{-1em}
\begin{equation}
\max_{\alpha}{\enskip Q(\alpha)} \quad \text{subject to} \quad T(\alpha) \approx T_0
\label{eq:equality-constrained-opt}
\end{equation}
\vspace{-1.3em}

\noindent The above can be relaxed as an unconstrained optimization problem that aims to maximize

\vspace{-1em}
\begin{equation*}
    r(\alpha) = Q(\alpha) + \beta \left|T(\alpha) / T_0 - 1\right|
\end{equation*}
\vspace{-1.5em}

\noindent where $|\cdot|$ denotes the absolute function and $\beta < 0$, the \emph{cost exponent}, is a finite negative scalar that controls how much strongly we encourage architectures to have inference times close to $T_0$. The expression $T(\alpha) / T_0$ ensures the reward is scale-invariant w.r.t.\ latency. Search results are robust to the exact value of $\beta$,\footnote{We used the same value of $\beta$ for all our classification experiments.} and this scale-invariance further reduces the need to retune $\beta$ for new devices and search spaces.

Using the absolute value reward, we found that while the \emph{average} inference time of models sampled by the RL controller was consistently close to the target, the inference time of the \emph{most likely} architecture selected at the end of the search could be several milliseconds lower. We combat the mismatch between \emph{average} and \emph{most likely} inference times by adjusting the learning rate schedule of the RL controller. Instead of using a constant learning rate through the search, we exponentially \textbf{increase} the RL learning rate over time. This allows the controller to explore the search space (with a relatively low learning rate) at the start of the search, but also ensures that the entropy of the RL controller is low at the end of the search, preventing the mismatch. Additional information is provided in Append \ref{sec:abs-reward-avg-vs-argmax}.

\subsection{Improving the shared model weights}
We identified two techniques that allowed us to improve the quality of the models found by architecture search experiments. Both techniques rely on the intuition that if we can ensure that all of our shared model weights are sufficiently well-trained, we can get a more reliable signal about which parts of the search space are most promising.

\vspace{1em}
\noindent \textbf{Filter Warmup.} We can efficiently search over different filter sizes by masking out tensors across the \emph{filters} dimension. For example, we can simulate a convolution with 96 output filters by taking a convolution with 128 output filters and zeroing out the rightmost 32. However, this introduces a bias into our training process: the left-most filters will always be trained, whereas the right-most filters will be trained only occasionally. To counteract this effect, we randomly enable \emph{all} the output filters -- rather than just the filters selected by the RL controller -- with some probability $p$. We linearly decrease $p$ from 1 to 0 over the first 25\% of the search,\footnote{We also experimented with decreasing $p$ over 50\% of steps instead of 25\%, but did not observe a significant effect on search quality.} during which the RL controller is disabled and only the shared model weights are trained.

\vspace{1em}
\noindent \textbf{Op Warmup.} We enable all possible operations in the search space at start of training, and gradually drop out more and more of the operations as the search progresses. Our discussion will focus on a single \emph{choice block}, where the goal is to select one of N possible operations (e.g., convolutions, inverted bottleneck layers, etc.) from a predetermined search space. The idea was originally proposed and evaluated by Bender et al. \cite{bender2018understanding}, who used carefully-tuned heuristics to determine the dropout schedule. We found that a simplified version of the idea could be used to improve our search results with minimal tuning. With some probability $p$ between 0 and 1 we enable \emph{all} operations within a choice block, rather than just enabling the operations selected by the RL controller. When multiple operations are enabled, we average their outputs. When $p = 0$, the controller's behavior is unaffected by op warmup. When $p = 1$, we enable all possible operations at every training step. In our implementation, we linearly decrease $p$ from 1 to 0 over the first 25\% of the search, during which the RL controller is disabled and only the shared model weights are trained. Op warmup can be implemented in a memory-efficient manner by leveraging rematerialization (Appendix \ref{ref:rematerialization}).

\section{Experimental Setup and Baselines}

For the ProxylessNAS and ProxylessNAS-Enlarged search spaces, our searches target the same latency as the published ProxylessNAS Mobile architecture \cite{cai2018proxylessnas}. For our MobileNetV3-Like search space, we target the latency of MobileNetV3-Large \cite{howard2019searching}. The resulting architectures are trained from scratch to obtain validation and test set accuracies. Unless otherwise specified, we obtained validation set accuracies of standalone models on a held-out subset of the ImageNet training set, after training on the remainder for 90 epochs. We obtained test set accuracies after training on the full training set for 360 epochs. Details for standalone training and architecture search are reported in Appendix \ref{sec:experimental-setup}.

For architecture search experiments, we always repeat the entire \emph{search process} 5 times as suggested by Lindauer and Hutter \cite{lindauer2019best}, and report the mean and variance of the performance of the resulting architectures. This is different from the popular practice of training a single resulting architecture multiple times, which only reflects the variance of a single searched architecture (which can be cherry-picked) rather than the search algorithm itself. For reproductions of published models such as \mobilenetv{2} where no architecture search is required on our part, we report the mean and variance across five training runs.

Identical hyper-parameters are used to train all of our models. (See Appendix \ref{sec:experimental-setup} for details.) The one exception is the dropout rate of the final fully connected layer, which is set to 0 when training for 90 epochs, 0.15 when training \mobilenetv{2}-based models for 360 epochs, and 0.25 when training \mobilenetv{3}-based models for 360 epochs. We initially experimented with tuning hyper-parameters such as the learning rate and dropout rate separately for each model. However, this did not lead to any additional quality gains, and was omitted for our final experiments.

\subsection{Reproducing Reference Architectures}
We start by reproducing three popular mobile image classification models in our training setup: \mobilenetv{2} \cite{sandler2018mobilenetv2}, \mnasnet{}-B1 \cite{tan2019mnasnet}, and \proxylessnas~\cite{cai2018proxylessnas}. This serves two purposes. First, it allows us to verify our training and evaluation setup. And second, it allows us to cleanly distinguish between improvements in our model training setup and improvements in our searched network architectures.

Results are presented in Table \ref{table:baseline-model-repro}. Our hyper-parameters provide quality comparable to the published results for \mnasnet{}-B1 and significantly improve upon the published results of \proxylessnas{} and \mobilenetv{2}.

There is an especially large (1.3\%) accuracy improvement in our reproduction of \mobilenetv{2}. This suggests that some of the quality gains from \mnasnet{} and \proxylessnas{} which were previously attributed to better network architectures may in fact be due to better hyper-parameter tuning. It underscores the importance of accounting for differences in training and evaluation setups when comparing different network architectures.

\begin{table}[t]
\centering
{\scriptsize

        \begin{tabular}{c|cccc}
        \toprule
        Name & Simulated & \multicolumn{3}{c}{Accuracy (\%)} \\ \cline{3-5}
        & Latency & Valid & Test & Test \\
        & & ours & published & ours \\
        \midrule
        \mobilenetv{2} & 77.2 ms & 74.5 $\pm$ 0.1 & 72.0 & \textbf{73.3 $\pm$ 0.1} \\
        \mnasnet{}-B1 & 84.5 ms & 76.0 $\pm$ 0.1 & \textbf{74.5} & \textbf{74.5 $\pm$ 0.1} \\
        \proxylessnas{} & 84.4 ms & 76.3 $\pm$ 0.2 & 74.6 & \textbf{74.9 $\pm$ 0.1} \\
        \midrule
        \mobilenetv{3} & 58.5 ms & 76.5 $\pm$ 0.2 & 75.2 & \textbf{75.3 $\pm$ 0.1} \\
        \bottomrule
        \end{tabular}
        }
    \caption{Reproductions of our baseline models on ImageNet.}
    \label{table:baseline-model-repro}
    
\vspace{1em}
{\scriptsize
    \begin{tabular}{@{}c|ccc@{}}
\toprule
Model / Method & Valid Acc (\%) & Test Acc (\%) & Latency \\ \midrule
ProxylessNAS \cite{cai2018proxylessnas} & 76.2 & 74.8 & 84.4 \\
\midrule
RS ($N=20$) & 75.4 $\pm$ 0.2 & 73.9 $\pm$ 0.3 & 84.3 $\pm$ 0.8 \\
RS ($N=50$) & 75.4 $\pm$ 0.2 & 74.0 $\pm$ 0.2 & 83.8 $\pm$ 0.6 \\
TuNAS (90 epochs) & 76.3 $\pm$ 0.2 & \textbf{75.0 $\pm$ 0.1} & 84.0 $\pm$ 0.4 \\ \bottomrule
\end{tabular}
}
\caption{Results in the ProxylessNAS search space. ``ProxylessNAS \cite{cai2018proxylessnas}'' is our reproduction of the ProxylessNAS-Mobile model. Our TuNAS implementation includes op/filter warmup, the absolute value reward, and more aggressive weight sharing.}
\label{tab:proxylessnas-search-results}

\vspace{1em}
{\scriptsize

\begin{tabular}{@{}c|ccc@{}}
\toprule
Model / Method & Valid Acc (\%) & Test Acc (\%) & Latency \\
\midrule
MobileNetV2 \cite{sandler2018mobilenetv2} & 74.4 & 73.4 & 77.2 \\
MNASNet-B1 \cite{tan2019mnasnet} & 76.0 & 74.5 & 84.5 \\
ProxylessNAS \cite{cai2018proxylessnas} & 76.2 & 74.8 & 84.4 \\
\midrule
RS ($N=20$) & 74.4 $\pm$ 0.5 & 73.1 $\pm$ 0.6 & 84.0 $\pm$ 0.6 \\
RS ($N=50$) & 74.6 $\pm$ 0.3 & 73.2 $\pm$ 0.3 & 83.5 $\pm$ 0.3 \\
TuNAS (90 epochs) & 76.4 $\pm$ 0.1 & \textbf{75.3 $\pm$ 0.2} & 84.0 $\pm$ 0.4 \\
\bottomrule
\end{tabular}
}
\caption{Results in the \proxylessnas{}-Enlarged search space.}
\label{tab:proxylessnas-enlarged-search-results}

\vspace{1em}
{\scriptsize
\begin{tabular}{@{}c|ccc@{}}
\toprule
Model / Method & Valid Acc (\%) & Test Acc (\%) & Latency \\ \midrule
MobileNetV3-L \cite{howard2019searching} & 76.5 & 75.3 & 58.5 \\ \midrule
RS ($N=20$) & 74.1 $\pm$ 0.6 & 73.0 $\pm$ 0.5 & 58.5 $\pm$ 0.5 \\
RS ($N=50$) & 74.6 $\pm$ 0.3 & 73.5 $\pm$ 0.2 & 58.7 $\pm$ 0.4 \\
TuNAS (90 epochs) & 76.6 $\pm$ 0.1 & 75.2 $\pm$ 0.2 & 57.0 $\pm$ 0.2 \\
TuNAS (360 epochs) & 76.7 $\pm$ 0.2 & \textbf{75.4 $\pm$ 0.1} & 57.1 $\pm$ 0.1 \\
\bottomrule
\end{tabular}
}
\caption{Results in the MobileNetV3-Like search space.}
\label{tab:mobilenetv3-like-search-results}
\end{table}

\section{Results and Discussion}
Compared with previous papers on efficient architecture search such as \proxylessnas{}, our architecture search setup includes several novel features, including (i) a new absolute value reward, (ii) the use of op and filter warmup, and (iii) more aggressive weight sharing during searches. At the end of this section we will systematically evaluate these changes.
First we evaluate our proposed efficient architecture search implementation on the three search spaces presented in Section \ref{sec:search-spaces}, and compare our results against random search with similar or higher search cost. The search spaces gradually increase in terms of both size and difficulty.

\vspace{1em}

\noindent \textbf{Finding 1: TuNAS outperforms Random Search (RS) by a large margin in each of our three classification search spaces. This holds even though we use 2-3x more compute resources for each RS experiment (Appendix \ref{sec:random-search-cost}).}

First (Table \ref{tab:proxylessnas-search-results}), we evaluate our search algorithm on the \proxylessnas{} search space~\cite{cai2018proxylessnas}. Despite having lower search costs, the accuracies of architectures found with an efficient architecture search improve upon random search by 1\%. These also provide a sanity check for our setup: the results of our search are competitive with those reported by the original \proxylessnas{} paper.

Next (Table \ref{tab:proxylessnas-enlarged-search-results}), we evaluate our search algorithm on the \proxylessnas{}-Enlarged search space, which additionally searches over output filter sizes. In this larger and more challenging search space, the accuracy gap between our method and random search increases from 1\% to 2\%.

Finally (Table \ref{tab:mobilenetv3-like-search-results}), we evaluate our search algorithm on our \mobilenetv{3}-Like search space, which is the most challenging of the three. In addition to being the largest of the three search spaces, the accuracies of architectures sampled from this search space are -- on average -- lower than the other two. Increasing the size of a random sample from $N=20$ to $N=50$ can improve the results of random search. However, we find that increasing the search time of our algorithm from 90 to 360 epochs can also improve the results of our efficient algorithm, while still maintaining a lower search cost than random search with $N=50$.

\vspace{1em}
\noindent \textbf{Finding 2: The TuNAS implementation generalizes to object detection.} We investigate the transferability of our algorithm to the object detection task,
by searching for the detection backbone w.r.t.\ both mean average precision and the inference cost.
Results on COCO~\cite{lin2014microsoft} are summarized in Table~\ref{tab:coco-results}. The searched architecture outperforms the state-of-the-art model MobileNetV3 + SSDLite \cite{howard2019searching}.
Details of the experimental setup are presented in Appendix \ref{sec:object-detection-setup}.

\begin{table}
\centering

{\scriptsize
\centering
\begin{tabular}{@{}c|cc@{}}
\toprule
Backbone & COCO Test-dev mAP & Latency \\ \midrule
MobileNetV2 & 20.7 & 126 \\
MNASNet & 21.3 & 129 \\
ProxylessNAS & 21.8 & 140 \\
MobileNetV3-Large & 22.0 & 106 \\ \midrule
TuNAS Search & \textbf{22.5} & 106 \\ \bottomrule
\end{tabular}
}
\caption{Backbone architecture search results on MS COCO in the MobileNetV3-Like space. All detection backbones are combined with the SSDLite head. Target latency for TuNAS search was set to 106ms (same as for MobileNetV3-Large + SSDLite).}
\label{tab:coco-results}
\vspace{1em}
{\scriptsize

        \begin{tabular}{c|ccc}
            \toprule
            Filters & Valid Acc (\%) & Test Acc (\%) & Latency \\ \midrule
            ProxylessNAS & 76.3 $\pm$ 0.2 & \textbf{75.0 $\pm$ 0.1} & 84.0 $\pm$ 0.4 \\
            $\times$2 Every Stride-2 & 74.8 $\pm$ 0.2 & 73.5 $\pm$ 0.2 & 83.9 $\pm$ 1.0 \\
            $\times$2 Every Block & 75.3 $\pm$ 0.2 & 74.0 $\pm$ 0.2 & 83.9 $\pm$ 0.2 \\
            \bottomrule
        \end{tabular}
        }
    \caption{Effect of output filter sizes on final model accuracies.}
    \label{tab:output-filter-sizes}
\vspace{1em}
{\scriptsize

        \begin{tabular}{c|ccc}
            \toprule
            Reward & Valid Acc (\%) & Test Acc (\%) & Latency \\
            \midrule
            \mnasnet{}-Soft Reward & 76.2 $\pm$ 0.2 & 74.8 $\pm$ 0.3 & 79.5 $\pm$ 3.3 \\
            Absolute Value Reward & 76.4 $\pm$ 0.1 & 75.0 $\pm$ 0.1 & 84.1 $\pm$ 0.4 \\
            \bottomrule
        \end{tabular}
        }
    \caption{Comparison of our absolute value reward function ($T_0$=84ms) vs. the reward used by \mnasnet{} and \proxylessnas{}. While both provide similar quality/latency tradeoffs, our absolute value reward allows precise control over the inference latency, and reduces the need for extra tuning to find a suitable cost exponent.}
    \label{tab:proxylessnas-soft-vs-abs-reward-search}
\vspace{1em}
{\scriptsize
        \begin{tabular}{cc|ccc}
            \toprule
            Search Space & Warmup & Valid Acc (\%) & Test Acc (\%) & Latency \\ \midrule
            \proxylessnas{} & \ding{55} & 76.1 $\pm$ 0.1 & 74.7 $\pm$ 0.1 & 84.0 $\pm$ 0.3 \\
            \proxylessnas{} & \ding{51} & 76.3 $\pm$ 0.2 & \textbf{75.0 $\pm$ 0.1} & 84.0 $\pm$ 0.4 \\
            \midrule
            \proxylessnas{}-Enl & \ding{55} & 75.8 $\pm$ 0.3 & 74.6 $\pm$ 0.2 & 83.6 $\pm$ 0.2 \\
            \proxylessnas{}-Enl & \ding{51} & 76.4 $\pm$ 0.1 & \textbf{75.3 $\pm$ 0.2} & 84.0 $\pm$ 0.4 \\
            \midrule
            \mobilenetv{3}-Like & \ding{55} & 76.2 $\pm$ 0.2 & 75.0 $\pm$ 0.1 & 57.0 $\pm$ 0.6 \\
            \mobilenetv{3}-Like & \ding{51} & 76.6 $\pm$ 0.1 & \textbf{75.2 $\pm$ 0.2} & 57.0 $\pm$ 0.2 \\
            \bottomrule
        \end{tabular}
        }
    \caption{Comparison of search results with vs.\ without op and filter warmup. We use aggressive weight sharing and search for 90 epochs in all search configurations.}
    \label{tab:warmup-ablation}
\end{table}

\vspace{1em}
\noindent \textbf{Finding 3: Output filter sizes are important.}

\mnasnet{}-B1~\cite{tan2019mnasnet} searches over the number of output filters in addition to factors such as the kernel sizes and expansion factors. This is different from  many recent papers on efficient NAS--including ENAS ~\cite{pham2018efficient}, DARTS~\cite{liu2018darts}, and \proxylessnas{}~\cite{cai2018proxylessnas}--which hard-coded the output filter sizes.

To determine the importance of output filter sizes,
one possibility would be to modify the output filter sizes of a high-performing model (such as the ProxylessNAS-Mobile model) and look at how the model accuracy changes. However, we can potentially do better by searching for a \emph{new} architecture whose operations are better-adapted to the new output filter sizes. We therefore perform two different variants of the latter procedure. In the first variant, we replace the ProxylessNAS output filter sizes (which are hard-coded to be almost the same as MnasNet) with a naive heuristic where we double the number of filters whenever we halve the image width and height, similar to architectures such as ResNet and VGG. In the second, we double the number of filters at each new block. Table \ref{tab:output-filter-sizes} shows that searched filter sizes significantly outperform both doubling heuristics.

\vspace{1em}
\noindent \textbf{Finding 4: Aggressive weight sharing enables larger search spaces without significant quality impact.}
We share weights between candidate networks more aggressively than previous works such as \proxylessnas{} (Section \ref{sec:aggressive-weight-sharing}). This lets us explore much larger search spaces, including one with up to 252 options per inverted bottleneck. For the \proxylessnas{} space (where searches are possible with and without it), we verified that it does not significantly affect searched model quality (Appendix \ref{sec:aggressive-weight-sharing-experiments}).

\vspace{1em}
\noindent
\textbf{Finding 5: The absolute value reward reduces hyper-parameter tuning significantly.} With the \mnasnet{}-Soft reward function, we found it necessary to grid search over $\beta \in \{ -0.03, -0.04, -0.05, -0.06, -0.07, -0.08, -0.09 \}$ in order to reliably find network architectures close to the target latency. %
By switching to the absolute value reward function, we were able to eliminate the need for this search, reducing resource costs by a factor of 7. We compared the quality of both methods on our implementation of the \proxylessnas{} search space with weight sharing, and found that the Absolute Value reward function did not significantly affect the quality/latency tradeoff (Table \ref{tab:proxylessnas-soft-vs-abs-reward-search} and Appendix \ref{sec:abs-value-reward-discussion}). %

\vspace{1em}
\noindent \textbf{Finding 6: Op and filter warmup lead to consistent improvements across all search spaces.}
Controlled experiments are presented in Table \ref{tab:warmup-ablation}. While improvements are small in some spaces, they account for nearly half of all quality gains in the ProxylessNAS-Enlarged Space.

\section*{Acknowledgements}
We would like to thank
Blake Hechtman,
Ryan Sepassi,
and
Tong Shen
for their help with software changes needed to make our algorithms to work well on TPUs. We would also like to thank
Berkin Akin,
Okan Arikan,
Yiming Chen,
Zhifeng Chen,
Frank Chu,
Ekin D. Cubuk,
Matthieu Devin,
Suyog Gupta,
Andrew Howard,
Da Huang,
Adam Kraft,
Peisheng Li,
Yifeng Lu,
Ruoming Pang,
Daiyi Peng,
Mark Sandler,
Yonghui Wu,
Zhinan Xu,
Xin Zhou,
and
Menglong Zhu
for helpful feedback and discussions.

{\small
\bibliographystyle{ieee_fullname}
\bibliography{main}
}
\cleardoublepage
\appendix

\section{Variance of \proxylessnas{}-Mobile Model}
\label{sec:variance}
For stand-alone model training, we estimated the variance in accuracy across runs by starting five identical runs of the \proxylessnas{}-Mobile model. We repeated the experiment in two configurations. In the first configuration, we trained for 90 epochs and then evaluated the models on our validation set; the resulting validation set accuracies were $[76.1\%, 76.4\%, 76.4\%, 76.5\%, 76.2\%]$. In the second configuration, we trained for 360 epochs and then evaluated on our test set; the resulting test set accuracies were $[75.0\%, 75.0\%, 74.9\%, 74.9\%, 75.0\%]$.\footnote{Accuracies on the validation set are typically a few percentage points higher than on the test set.}

\section{Average vs Argmax Inference Time}  \label{sec:abs-reward-avg-vs-argmax}
For the absolute value reward function with a constant RL learning rate, we argued that the reason models didn't converge to the target inference time was because of differences between the inference times of \emph{randomly sampled} models vs. the \emph{argmax} model taken by selecting the most likely choice for each categorical decision. To test this hypothesis, we compared the two at the end of a search. We obtained an average time from randomly sampled models using an exponential moving average with a decay rate of 0.9 which was updated every 100 training steps. While average times consistently converged to the desired inference time target, the inference time from the argmax could differ by 8 ms or more.

\begin{table}[h]
    \centering
    \begin{tabular}{c|c|c}
        $\beta$ & Avg. Time & ArgMax Time \\
        \hline
        -0.01 & 134.2 ms & 128.8 ms \\
        -0.02 & 104.3 ms & 92.2 ms \\
        -0.05 & 84.2 ms & 75.8 ms \\
        -0.10 & 83.1 ms & 78.3 ms \\
        -0.20 & 84.0 ms & 82.7 ms \\
        -0.50 & 84.2 ms & 83.4 ms \\
        -1.00 & 83.2 ms & 85.5 ms
    \end{tabular}
    \caption{Average vs. argmax inference times when using the absolute value reward function without a learning rate schedule for the RL controller.}
    \label{tab:average-vs-most-likely-inference-time}
\end{table}

\section{Rematerialization}
\label{ref:rematerialization}
If op warmup is implemented naively then the activation memory required to train the shared model weights grows linearly with the number of possible operations in the search space. If many different possible operations can be simultaneously enabled at each position in the network, the model will be unable to fit in memory. We use rematerialization to address this issue. During the forward pass, we apply $N$ different operations to the same input, and average their outputs. Rather than retaining the intermediate results of these operations for use during the backwards pass, we throw them away. During the backwards pass, we recompute the intermediate results for one operation at a time. In practice, this leads to a large decrease in memory requirements, as we only need to retain a single input tensor and a single output tensor for each choice block. For example, in our reproduction of the original ProxylessNAS search space with a per-core batch size of 128, rematerialization decreases the memory needed to train a reproduction of the original \proxylessnas{} search space from 29.5 GiB to 4.8 GiB. This memory-saving technique, which allowed us to scale to larger search spaces, came at the cost of roughly a 30\% increase in search times to perform a second forward pass for each of the N possible operations.

Finally, we note that although this rematerialization trick was developed with our version of op rampup in mind, it could also be used to reduce the memory requirements of a method such as DARTS \cite{liu2018darts} which requires us to evaluate every possible operation in the search space at every training step.

\section{Discussion of Absolute Value Reward}
\label{sec:abs-value-reward-discussion}
We now contrast a typical architecture search workflow with the \mnasnet{}-Soft reward function against a workflow with our new Absolute Value reward function. For the \mnasnet{}-Soft reward function, the first step when using a new search space or training configuration is to tune the RL controller's cost exponent $\beta$ to obtain inference times which are reasonably close to our target latency. In our early experiments, we found that grid searching over $\beta \in \{ -0.03, -0.04, -0.05, -0.06, -0.07, -0.08, -0.09\}$ worked well in practice. However, running this grid search increased the cost of architecture search experiments by a factor of 7.

Even after we fixed the value of $\beta$, the latencies and accuracies of architectures found by a search could vary significantly from one run to the next. For example, in our reproduction of the \proxylessnas{} search space with $\beta = -0.07$, five identical architecture search experiments returned latencies which ranged from 74ms to 82ms. We also saw a wide variance in accuracies across the different architectures, ranging from 75.8\% to 76.4\% on the validation set and 74.2\% to 75.1\% on the test set. Larger models generally had better accuracies, indicating that the problem stemmed from our inability to precisely control the latency.

This helped motivate our Absolute Reward function, which allowed the RL controller to reliably find architectures whose latencies were close to the target. For example, the low variance of searched TuNAS model latencies in Tables \ref{tab:proxylessnas-search-results}, \ref{tab:proxylessnas-enlarged-search-results}, \ref{tab:mobilenetv3-like-search-results}, and \ref{tab:aggressive-weight-sharing-search} shows we can reliably find high-quality architectures within 1 ms of the target across several different search configurations, even when we reuse the same search hyper-parameters between different setups. %

As an alternative to the absolute value reward function, we also considered searching for an architecture close to the inference time target, and then uniformly scaling up or down the number of filters in every layer. While this helped reduce the variance of searched model accuracies, it did not remove the need to tune the RL cost exponent, since we needed to find a model that was already close to the inference time target to get good results. Furthermore, finding the right scaling factor to hit a specific inference time target added an extra step to experiments in this setup. The absolute value reward function gave us high-quality architectures with a more streamlined search process. 

\section{Experimental Setup}
\label{sec:experimental-setup}
\subsection{Standalone Training for Classification}

During stand-alone model training, each model was trained using distributed synchronous SGD on TensorFlow with a Cloud TPU v2-32 or Cloud TPU v3-32 instance (32 TPU cores) and a per-core batch size of 128. Models were optimized using RMSProp with momentum = 0.9, decay rate = 0.9, and epsilon = 0.1. The learning rate was annealed following a cosine decay schedule without restarts \cite{loshchilov2016sgdr}, with a maximum value of 2.64 globally (or 0.0825 per core). We linearly increased the learning rate from 0 over the first 2.5\% of training steps \cite{goyal2017accurate}. Models were trained with batch normalization with epsilon = 0.001 and momentum = 0.99. Convolutional kernels were initialized with He initialization \cite{he2016deep},\footnote{TensorFlow's default variable initialization heuristics, such as \texttt{tf.initializers.he\_normal} are designed for ordinary convolutions, and can overestimate the fan-in of depthwise convolutional kernels by multiple orders of magnitude; we corrected this issue in our version.} while bias variables were initialized to 0. The final fully connected layer of the network was initialized from a random normal distribution with mean 0 and standard deviation 0.01. We applied L2 regularization with a strength of 0.00004 to all convolutional kernels except  the final fully connected layer of the network. All models were trained with ResNet data preprocessing and an input image size of 224$\times$224 pixels. When training models for 360 epochs, we applied a dropout rate of 0.15 before the final fully connected layer for models from \mobilenetv{2} search spaces and 0.25 for models from \mobilenetv{3} search spaces. We did not apply dropout when training models for 90 epochs.

As is standard for ImageNet experiments, our test set accuracies were obtained on what is confusingly called the ImageNet validation set for historical reasons. What we refer to as validation set accuracies were obtained on a held-out subset of the ImageNet training set containing 50,046 randomly selected examples.

\subsection{Architecture Search for Classification} \label{sec:architecture-search-setup}
Architecture search experiments are performed using Cloud TPU v2-32 or Cloud TPU v3-32 instances with 32 TPU cores and a per-core batch size of 128.

For training the shared model weights, we use the same hyper-parameters as for stand-alone model training, except that the dropout rate of the final fully connected layer is always set to 0. When applying L2 regularization to the traininable model variables, we only regularize parameters which are used in the current training step. Because batch norm statistics can potentially vary significantly from one candidate architecture to the next, batch norm is always applied in ``training'' mode, even during model evaluation.

For training the RL controller, we use an Adam optimizer with a base learning rate of 3e-4, $\beta_1 = 0$, $\beta_2 = 0.999$, and $\epsilon = $1e-8. We set the learning rate of the RL controller to 0 for the first 25\% of training. If using an exponential schedule, we set the learning rate equal to the base value 25\% of the way through training, and increase it exponentially so that the final learning rate is 10x the base learning rate. If using a constant schedule, we set the learning rate equal to the base learning rate after the first 25\% of training.

\subsection{Object Detection}
\label{sec:object-detection-setup}
Our implementation is based on the Tensorflow Object Detection API \cite{huang2017speed}.
All backbones are combined with SSDLite \cite{sandler2018mobilenetv2} as the head. Following MobileNetV2 \cite{sandler2018mobilenetv2} and V3 \cite{howard2019searching}, we use the last feature extractor layers that have an output stride of 16 (C4) and 32 (C5) as the endpoints for the head. In contrast with MobileNetV3 + SSDLite \cite{howard2019searching}, we do \emph{not} manually halve the number of channels for blocks between C4 and C5, since in our case the number of channels is automatically learned by the search algorithm.
All experiments use 320$\times$320 input images.

For standalone training, each detection model is trained for 50K steps from scratch on COCO train2017 data using a Cloud TPU v2-32 or TPU v3-32 instance (32 TPU cores) with a per-core batch size of 32. We use SGD to optimize the shared model weights with a momentum of 0.9. The (global) learning rate is warmed up linearly from 0 to 4 during the first 5K steps and then decayed to 0 following a cosine schedule \cite{loshchilov2016sgdr} during the rest of the training process.

For architecture searches, training configurations for the model weights remain the same as for standalone training. We split out 10\% of the data from COCO train2017 to compute the reward during an architecture search. The training setup of the RL controller is the same as for classification, except that the base learning rate of the Adam optimizer is set to 5e-3. Whereas classification models are evaluated based on accuracy, detection models are evaluated using mAP (mean Average Precision).
To obtain results in Table \ref{tab:coco-results},
architecture searches were carried out in the MobileNetV3-Like search space with a target inference cost of 106ms to match the simulated latency of MobileNetV3 + SSDLite.

To obtain the test-dev results, each model is trained over the combined COCO train2017 and val2017 data for 100K steps. Other settings remain the same as those for standalone training and validation.

\subsection{Simulated Inference Times}
In early experiments, we found that if we benchmarked the same model on two different phones, the observed latencies could differ by several milliseconds. To ensure that our results were reproducible -- and to mitigate the possibility of random hardware-specific variance across runs -- we estimated the latencies of our models using lookup tables similar to those proposed by NetAdapt \cite{yang2018netadapt}. These lookup tables let us estimate the latency of each individual operation (e.g., convolution or pooling layer) in the network. The overall latency of a network architecture was estimated by summing up the latencies of all its individual operations.

\begin{figure}
    \centering
    \includegraphics[width=0.6\linewidth]{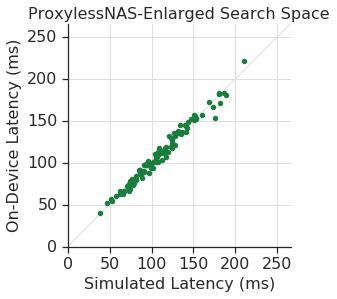}
    \includegraphics[width=0.6\linewidth]{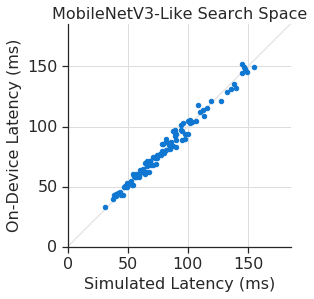}
    \caption{On-device vs. simulated latencies in the ProxylessNAS-Enlarged and MobileNetV3-Like search spaces. Each plot is based on 100 random architectures in the given space.}
    \label{fig:latency-calibration}
\end{figure}

We validated our use of simulated latencies by sampling 100 random architectures and comparing the simulated numbers against on-device numbers measured on a real Pixel-1 phone. Figure \ref{fig:latency-calibration} shows that the two are well-correlated.

\section{Cost of Random Search vs Efficient NAS}
\label{sec:random-search-cost}

Training a single architecture for 90 epochs on ImageNet requires about 1.7 hours using a Cloud TPU v3-32 instance (32 cores), whereas a single architecture search run takes between 8 and 13 hours, depending on the search space. This means that for the cost of a single 90-epoch search, we can evaluate 4-8 random models. In some cases, we found that the cost of an efficient architecture search could be further improved by increasing the number of search epochs from 90 to 360. For the cost of a single 360-epoch search, we can evaluate 15 - 30 random models. We provide a generous budget of 20 - 50 models for our random search experiments in order to demonstrate that efficient architecture search can outperform random search even if each random search experiment is more compute-intensive.

\section{Quality of Aggressive Weight Sharing} \label{sec:aggressive-weight-sharing-experiments}
To verify the quality impact of aggressive weight sharing, we ran architecture searches on the original \proxylessnas{} search space both with and without aggressive sharing. The results (Table \ref{tab:aggressive-weight-sharing-search}) indicate that aggressive weight sharing does not significantly affect searched model accuracies in this space. Our other two search spaces (\proxylessnas{}-Enlarged and MobilenetV3-Like) were too large us to run searches without aggressive weight sharing.

\begin{table}[tb]
        \centering
        \footnotesize
        \begin{tabular}{c|ccc}
        \toprule
            Agg. Sharing & Valid Acc (\%) & Test Acc (\%) & Latency \\ \midrule
            No & 76.4 $\pm$ 0.1 & \textbf{75.0 $\pm$ 0.1} & 84.1 $\pm$ 0.4 \\
            Yes & 76.3 $\pm$ 0.2 & \textbf{75.0 $\pm$ 0.1} & 84.0 $\pm$ 0.4 \\
            \bottomrule
        \end{tabular}
    \caption{Effect of aggressive weight sharing (abbreviated as ``Agg Sharing'' in the table above) on the quality of searched architectures. Each search is run for 90 epochs on the \proxylessnas{} search space with op and filter warmup enabled.}
    \label{tab:aggressive-weight-sharing-search}
\end{table}

\cleardoublepage
\end{document}